# The method of automatic summarization from different sources

*N. Shakhovska, T. Cherna*

**Information systems and networks department
Lviv Polytechnic National University,
S. Bandery str., 12, Lviv, 79013, Ukraine;
e-mail: natalya233@gmail.com**

*Abstract.* In this article is analyzed technology of automatic text abstracting and annotation. The role of annotation in automatic search and classification for different scientific articles is described. The algorithm of summarization of natural language documents using the concept of importance coefficients is developed. Such concept allows considering the peculiarity of subject areas and topics that could be found in different kinds of documents.

Method for generating abstracts of single document based on frequency analysis is developed. The recognition elements for unstructured text analysis are given. The method of pre-processing analysis of several documents is developed. This technique simultaneously considers both statistical approaches to abstracting and the importance of terms in a particular subject domain. The quality of generated abstract is evaluated. For the developed system there was conducted experts evaluation. It was held only for texts in Ukrainian. The developed system concluding essay has higher aggregate score on all criteria.

The summarization system architecture is building. To build an information system model there is used CASE-tool AllFusion ERwin Data Modeler. The database scheme for information saving was built.

The system is designed to work primarily with Ukrainian texts, which gives a significant advantage, since most modern systems still oriented to English texts.

*Key words:* annotation, abstracting, national system of abstracting, heterogeneous data, analysis.

## INTRODUCTION

Automatic abstracting (came from *Automatic Text Summarization* ) is a creation of summaries of materials, summaries or digests, that is getting of the most important data, from one or from several documents and generation on their basis laconic and informal sated reports. There are two directions of automatic summarization – quasi summarization and a maintenance summary.

## THE ANALYSIS OF RECENT RESEARCHES AND PUBLICATIONS

As a branch of practical activities summarization was generated in depths of library-bibliographic and journalistic practice. In the beginning of XX century the basic task of summarization is full reflection of the maintenance of the primary source on what the attention of experts – experts has been concentrated. Activation of development of theoretical questions of annotation was promoted by introduction of system of the annotated printing card in 1925 .During this period appears a lot of works where the concept of "summary" was considered, classification of summaries and general provisions concerning their drawing up was directed. Very fruitful was a worked out the scheme of the summary offered by O.G.Fomin [1]. He opened about 30 information elements which are used in texts of summaries, having allocated among them the main things homogeneous behind a design and the form.

With formation and development of computer science a new stages in development summarization were based, on the agenda the attention to the question of algorithmization and automation of this process was brought. Problems of summarization which gets an interdisciplinary character are studied by bibliographers, linguists, philosophers, experts in branch of computer science, computer facilities and cybernetics. In researches of 1970th years the attention was concentrated on principles of manual summarization and possibilities of formalization of the given process with computer facilities use. The great value is got by problems of automation of semantic processes which developed behind three basic directions. The first – has been connected with introduction of inquiries to system. This process requires the previous semantic transformation of inquiry that is construction on demand of the search instruction which is already entered to the information retrieval system. The second direction covered the search processes connected with comparison behind certain rules of search images of inquiries which have arrived to system, and to a documents which remain there. The third direction pursued the aim of automation to input documents to the information retrieval system. Among its processes the central place is occupied with questions of automation of indexing, annotation and summarization. More details about problem automatic summarization you can see on VINITI analytical review [1].

Statistical methods are based on the American scientific G.Lun's [7] workings out which the first in 1958 has received the machine abstract. He has suggested to carry out selection of offers on the basis of frequency of the use of words in the offer (this is more often, there is a word in it, the above its semantic weight), and also in view of the location of significant words in the offer. In selection of offers to the abstract for each of them "the semantic weight" is defined. Than more words which often meet, appear nearly, the more especially essential information is contained by the offer, as should join in the abstract.

G.Lun's ideas stimulated the subsequent workings out from automation summarization, based on the statistical analysis of texts. Techniques of Russians V.Agraev, B.Borodin [7] and V.Purto [11] became the most known. First two have offered a technique; agree with which offers chosen from the text appear connected among themselves. The most connected that is why such which are subject to inclusion to the abstract, offers which contain the greatest quantity of identical significant words are considered.

V.Purto, V.Gorkov [11] has developed a method of an estimation and selection of offers behind quantity of the information which they contain [7]. In this case texts are subject to the statistical analysis for revealing of frequency of the use of words. Words which are used in the scientific and technical literature more often, terms are. The researcher confirms: the more important term, the more often it meets in the text, and offers will contain a maximum quantity of these terms are selected. The volume of the abstract received in such a way makes, as a rule, no more than three offers, irrespective of volume of the primary document.

In case of use of a statistical method of summarization the volume and quality of abstracts completely depend on statistical characteristics of the text, therefore offers, which contain the major information (for example, conclusions in scientific articles, patent formulas in descriptions of inventions), can be not allocated absolutely not and not enter to the abstract. However, lacks are defined, to a certain extent, are compensated thanks to simplicity of the analysis and uniformity of abstracts which were prepared by means of the computer. It stimulates works in such direction in many countries.

Item methods are based on the direction of better work the most significant offers aimed at perfection in texts with attraction of a difficult mathematical apparatus. Selection is carried out on principles of four interconnection methods: a hint, keywords, the title, localization.

The essence of a method of a hint consists in use during selection of offers of the list of words in which words with positive, negative semantic weight and "zero" (neutral) are preliminary allocated. At selection words which transfer a positive and negative estimation are considered only. In case of use of a method of keywords the words selected by a frequency principle and to this sign defined key, there are considered in similar to the offered G.Lun [7] to the approach. In a method of the title the leading role is taken away to the dictionary of the terms selected from the title and subtitle which have big "weight", than words from other offers of the text. To the abstract offers where there are terms which are in the dictionary are selected. The localization method is based on the assumption, that the most essential information concentrates in the beginning or in the end of a certain fragment or the text paragraph. Comparison of all has shown four methods, that the method of keywords provides completeness of reflection of the maintenance of the primary document on 15-40 %, a title method – on 30-40 %, and general use of methods of a hint, the title and localization – on 30-60 %.

This approach has got the subsequent development during working out of indicative methods of summarization compared with which statistical and item methods play an auxiliary role. Indicative methods give the chance to formalize on the basis of parse a statement of the basic maintenance of the primary document in the abstract of cable style. To parse can be subject both all text, and its separate fragments which contain typical markers.

In the early eighties science officers of Moscow and Leningrad had been offered a technique formalized summarization with use of markers for texts from electronics. According to with this technique process automatic summarization is shown to two procedures:

own extracting (that is recognition in the text of the primary document of the marked offers and their deliveries on the printer); post editing (during it there is an elimination of logic and semantic repetitions, superfluous turns and also inclusions of necessary semantic sheaves between phrases). In this technique that post editing is carried out is positive is formalized and can be carried out not only experts in certain branch of knowledge, but also librarians, bibliographers who own bases of summarization. In the long term authors intended to realize the given technique so that the user independently defined a priority and made a choice of the aspects of the maintenance necessary to it (individual summarization) and, at necessity, a smog receive the corresponding abstract review of a file of primary sources [4].

The question of automation of semantic processes is considered and on pages of the Ukrainian professional editions. As a whole generalization of domestic and foreign experience testifies that the main aspects of research of problems of summarization are the history, the theory and a technique, and also the organization of processes of analytic-synthetic processing of the information.

SJ (summarization journal) were one of the very first forms scientifically information publications and within last century have gained the greatest distribution among the given kind of editions. And the abstract and the summary is one of the major means of scientific communications who function independently in separate systems (the scientific and technical information, libraries, publishing houses) and simultaneously carry out function of communication during an information transfer from one system to another. Theoretical and methodical aspects of analytic-synthetic processing of documents always were and remain in the centre of attention of experts in library business and scientifically to an information work [19, 20, 21].

Recently release of scientific and other editions not only in printing, but also the electronic form quickly extends [2, 8, 9]. Possibilities of carrying out of contextual search directly behind texts of publications have appeared and develop, that, apparently, does SJ superfluous. But such conclusion is erroneous. Information search behind abstracts which are collected in one file and in which in the compact form the basic maintenance of publications is stated, provides high revelation the received results while navigation behind full texts of publications in which there are almost all words of a natural language, causes considerable search noise. Therefore the subsequent development of computer technologies will lead not to dying off of the abstract information in the printing or electronic form, and to growth of its role, including as to navigation means in electronic libraries. Almost for 170 years of the existence the abstract information was well entered in system of scientific communications which has developed and for which for today there is no adequate replacement.

In view of continuous and fast growth of quantity of the printing and neo published information in the modern world and increases of the prices for a subscription, one of the most real and its valid means full coverage in the course of formation of an information DB are close cooperation of libraries and information centers from creation of the distributed information resources for mutual use. To solve these questions probably by creation of the big information systems for the purpose of fuller coverage subjects and a timely information transfer, elimination of unjustified duplication, the account of economic advantages from cooperation of libraries and information centers – participants of systems for analytic-synthetic processing and indexing of documents and so forth are created. Integration of information systems promotes increase of efficiency of an information work, its faster development and perfection [5, 6].

OBJECTIVES

Purpose of this paper is to improve the efficiency and quality of summarization method in text by the use of the combined algorithm and create own measure. This measure we can use for the sentence ordering method. This is the problem of taking several sentences, such as those deemed to be important by an extractive summarizer, and presenting them in the most coherent order. After that we will use text revision method. Sentence revision involves re-using text collected from the input to the summarizer, but parts of the final summary are automatically modified by substituting some expressions with other more appropriate expressions, given the context of the new summary. And the last method is sentence fusion. Sentence fusion is the task of taking two sentences that contain some overlapping information, but that also have fragments that are different. The goal is to produce a sentence that conveys the information that is common between the two sentences, or a single sentence that contains all information in the two sentences, but without redundancy.

THE MAIN RESULTS OF THE RESEARCH

We will analyze the document $D$:
$$D = \{T, K, A, M, L\}, \qquad (1)$$

where: $T$ is name, $K$ is the set of keywords, $A$ is the set of authors, $M$ is main part, $L$ is literature [3].

Defining elements of the document is based on the allocation of such text features [17-18]:

– location in the document;
– location of a paragraph (left, right, centered);
– type of writing (bold, italic, underline, normal);
– character recognition.

Based on these characteristics formed the basis of the rules of recognition elements of the document (Table 1).

**Table 1**. The rules of recognition elements of the document

| id | type | place | paragraph | alpha | symbols |
|----|------|-------|-----------|-------|---------|
| 1 | title | BEGIN | {Center;Right} | {Bold} | |
| 2 | author | BEGIN | {Center; Left} | | {By;©: (C) } |
| 3 | keyword | BEGIN | | | {Keyword; Keywords; Ключові слова; Ключевые слова} |
| 4 | main | CENTER | | | |
| 5 | literature | END | | | {Typical, Italic} |

To form essay there is stand out from the main part of the sentence.

Bulk, in turn, is divided into fragments by divisions and sections, introduced by the authors. It is believed that the sentences that appear in the introduction and conclusion, with higher informative value than a sentence with the middle of text.

*The sentence ordering method*

First of all, we introduce the concept of weight sentence.

The coefficient is defined as the location:

$$Location = \frac{1}{n \cdot m}, \quad (2)$$

where: $n = \overline{1..3}, m = \overline{1..3}$ – the place calls to the main part and paragraph respectively. Begin and end of text or paragraph estimated value of 1, the middle is as 3. Coefficient key phrase is determined by entering the sentence $U$ of elements of a set of significant sentences from $A$ membership function:

$$Cuephrase = m_A(U), \quad (3)$$

where: $A=\{«Conclusion», «In the end», «By the way»…\}$.

Index of statistical significance is formed on the basis of visiting sentence keywords specified by the author of the article:

$$Statterm = m_K(U). \quad (4)$$

The value added is defined as the presence of terms related words sentences that appear in the article's headline to the total number of words in a sentence (words) except for words whose length is less than 3 characters:

$$Addterm = \frac{word}{words}. \quad (5)$$

The weight of text block $U$ is:

$$Weight(U)= Location(U)+ Cuephrase(U)+ \\ +Statterm(U)+ Addterm(U). \quad (6)$$

*The text revision method*

So after studding all the documents it is necessary to accomplish the following: to exclude a statement with hit the consolidated data repository and to perform the final sorting sentences. For the bringing task of the final ranking factor "information novelty" there is used the following method:

1. Let we have two sets of sentences $B = \varnothing$ i $A = \{A_i | i = 1, 2, ..., N\}$, $N$ is count of sentences in text. For every sentence $A_i$ the usefulness $P(i)_i$ set $q_i$: $P(i)_i = q_i, i = 1, 2, ..., N$.

2. The sentences from set $A$ sort Descending $P(i)_i$.

3. If $A_i$ has the biggest $P(i)_i$, we take it in $B$. The usefulness for sentences in $A$ set s $P(i) = P(i)/kq_i$, where: $k > 0$ – factor clipping similar sentences.

4. Is $A$ empty? If NOT, go to 1.

*The sentence fusion method*

The next problem is information estimating from different sources [15- 16]. For semi-structured data type text file with a known format – dictionary data types defined formatting released the text of the formatting, copying its contents:

$$object \rightarrow Find(p_{firmattype}(s_{object}(Dic))),$$

for each object:
Selection
    ParagraphFormat.Alignment = Left (1, formattype),
    Font.type = Mid (formattype, 3, 1),
    Font.Caps = Right (formattype, 1),
    InStr (1,. Text, Right (formattype, 2),
Copy.Selection.

Sentence fusion is the method fro semantic network building. The algorithm of semantic network building consists of such steps:

- Find subject constants in dictionary $par \rightarrow Find(s_{par}(K))$;

- Text selection: par.Selection;

- Words selections situated behind subject constants: S=Right (.Text, Find(' ∨ ',' ∨ '.'));

- $Create(par, Cg')$;

- $Insert(Cg', S)$.

*The information system building*

To build an information system model there is used CASE-tool AllFusion ERwin Data Modeler, which enables model based infological model of information system build its datalogical model and create a database in any database management system. The development of the summarization system provides in the notation IDEF1X. The projected information system contains the following information objects (entities, IE) [10] (Fig. 1).

Sentence contains information about all the sentences in the text. The given entity has the following attributes:
- SentenceID serial number sentence;
- WordsWeight (Weight of words) is value of the sum of weights of all the words that make up sentences;
- Format is the sentence weight based on its format;
- Place is the sentence weight based on its location in the text;
- Sum is the sum of weights of the previous attribute that determines the total weight of the sentence.

*Keywords* contain information about all the key words in the text. The given entity has the following attributes:
- WordID contains the serial number of words;
- Word contains the word;
- Frequency is the weight of words, depending on the frequency of its appearance in the text;
- Place is the weight words depending on its location in the text;
- SentenceID is the order number sentence.Format is the weight words depending on its format;
- UserWeight is the weight of words defined by the user;
- Sum is the sum of the weights of previous attributes so that the total weight of the word.

Words-Sentence contains information about the relationship between words and sentences in the text. The given entity has the following attributes:
- ID is the serial number of communication;
- WordID is the serial number of words;

The main process in projected system "The abstract forming" divided into 5 subprocesses "Decode information from external format", "Wrapping text Structural branches", "Formation keywords", "Assigning weights", "Conclusions." These subprocesses performed in the system sequentially, one after the other. To perform "Decoding information from external format" subprocess on the entrance to it submitted data, which can change in the course of work. Such data are publication, the results of scientific research, scientific article. They enter the system as a result of user actions. The result of this execution and the input data according to "Wrapping text Structural branches" subprocess is a text information, ie data that entered the system after "Decoding information from external format" subprocess implementation are converted to the corresponding data format for the system in which they is ready for further processing. Due to performance "Wrapping text Structural branches" subprocess text information is divided into parts of a sentence. This subprocess occurs following the developed method. The result of this is a set of subprocess proposals submitted for subprocess entrance "Formation keywords" for further processing and for entrance "Assigning weights" subprocess. These subprocess are governed by rules of evaluation of proposals, and it uses a knowledge base. The result of "Assigning weights" subprocess is evaluated sentence. They arrive at the entrance to "Conclusions" subprocess given the results obtained as a result of the implementation of all previous subprocesses forms the abstract as the final result.

For the developed system there was conducted experts evaluation. It was held only for texts in Ukrainian. Method of estimation was as follows. Five experts were presented document the source and received on the basis of his essay. Experts answered the following questions by selecting the response under such scale evaluation [12-13]:
- To what extent abstract reflects the content of the documents? (1 – does not reflect, 2 – not quite complete, 3 – satisfactory).
- Is there redundancy in the abstract? (1-way too much, 2 – so not too much, 3 – no).
- Satisfies the abstract properties of connectivity text? (1 – or 2 – there are no related sentences 3 – true).
- Estimate the length of the essay (1 – very long, 2 – very short, 3 – the best).

The results of expert assessments are shown in Table 2.
- Most part of experts lowered estimates because there are sentences which violate the connectivity pattern of the text. The developed system concluding essay has higher aggregate score on all criteria. Highest rated experts put in the assessment of completeness, which means that the system is fairly accurate translation of the original document.

Developed automated system concluding essay can significantly reduce the time spent on drafting an essay in comparison with other systems summarization. Its algorithm is simple enough, but it has several advantages:
1. The use of weighting coefficients significantly improves the quality of the essay;
2. The user can determine the weight of certain terms, depending on what topic focused essay he wants to;
3. The system is designed to work primarily with Ukrainian texts, which gives a significant advantage, since most modern systems still oriented to English texts.

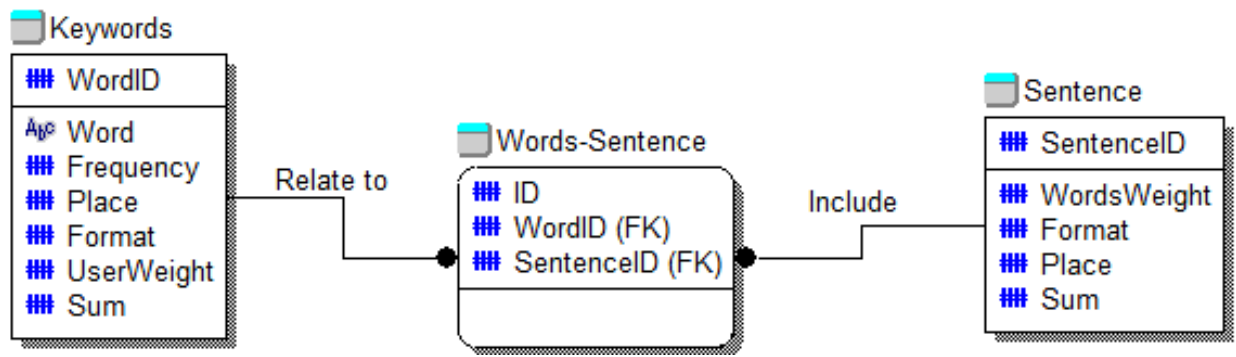

**Fig. 1**. Entity-relation diagram. The attribute level.

**Table 2**.Expert marks of essay quality

| The method | Expert number | Completeness | Redundancy | Connectivity Evaluation | Length Evaluation |
|---|---|---|---|---|---|
| Projected system | 1 | 2 | 2 | 2 | 3 |
|  | 2 | 3 | 2 | 1 | 1 |
|  | 3 | 3 | 3 | 3 | 2 |
|  | 4 | 2 | 1 | 2 | 3 |
|  | 5 | 2 | 3 | 2 | 3 |
|  | AVG | 2,35 | 2,05 | 1,89 | 2,22 |
| Autosummarize in Microsoft Office | 1 | 2 | 1 | 2 | 1 |
|  | 2 | 3 | 3 | 1 | 2 |
|  | 3 | 3 | 2 | 3 | 1 |
|  | 4 | 1 | 1 | 1 | 1 |
|  | 5 | 2 | 1 | 2 | 3 |
|  | AVG | 2,05 | 1,43 | 1,64 | 1,43 |

## CONCLUSIONS

1. The method of generating abstracts based on the developed models and methods of frequency analysis of terms in sentences and determining the weight of sentences was invented.

2. Summarizing of documents is an urgent and important task, which is very difficult for modern system of abstracting. Resources for referencing could have varied nature, so traditionally they required different pre-processing but using weights theory increases the quality of the resulting abstract. Methods of abstracting which are considered in this thesis can be successfully applied in modern information systems.